\documentclass{MITcsail}

\title{Designing deep neural networks for {driver} intention recognition}

\author
{Koen Vellenga~$^{1, 2}$, H. Joe Steinhauer~$^{1}$, Alexander Karlsson~$^{1}$, Göran Falkman~$^{1}$, Asli Rhodin~$^{3}$, Ashok Koppisetty~$^{2}$\\
\vspace{1em} 
\normalfont{\small $^{1}$University of Skövde, Sweden}\\
\normalfont{\small $^{2}$Volvo Car Corporation, Sweden}\\
\normalfont{\small $^{3}$Volvo Group, Sweden} \vspace{2em}
}

\usepackage{amssymb}
\usepackage{natbib}

\usepackage{lineno}
\usepackage{multirow}
\usepackage{longtable}
\usepackage{booktabs}
\usepackage{pgfplots}
\usepackage{pgfplotstable}
\usepackage{subcaption}
\newenvironment{conditions}
  {\par\vspace{\abovedisplayskip}\noindent
   \begin{tabular}{>{$}l<{$} @{} >{${}}c<{{}$} @{} l}}
  {\end{tabular}\par\vspace{\belowdisplayskip}}

\usepackage{hyperref} 

\begin{document}

\maketitle
\section*{Abstract}
\begin{abstract}
Driver intention recognition studies increasingly rely on deep neural networks. Deep neural networks have achieved top performance for many different tasks, but it is not a common practice to explicitly analyse the complexity and performance of the network's architecture. Therefore, this paper applies neural architecture search to investigate the effects of the deep neural network architecture on a real-world safety critical application with limited computational capabilities. We explore a pre-defined search space for three deep neural network layer types that are capable to handle sequential data (a long-short term memory, temporal convolution, and a time-series transformer layer), and the influence of different data fusion strategies on the driver intention recognition performance. A set of eight search strategies are evaluated for two driver intention recognition datasets. For the two datasets, we observed that there is no search strategy clearly sampling better deep neural network architectures. However, performing an architecture search does improve the model performance compared to the original manually designed networks. Furthermore, we observe no relation between increased model complexity and higher driver intention recognition performance. The result indicate that multiple architectures yield similar performance, regardless of the deep neural network layer type or fusion strategy.
\end{abstract}

\section{Introduction}
\label{sec:intro}
{Advanced driver assistance systems (ADAS) increasingly focus on actively supporting individual driving safety (e.g., forward collision warning, lane keeping assistance and lane departure warnings) (\cite{hasenjager2019survey}). The support of individual driver behavior is a challenging tasks, because human behavior can be irrational \cite{gabriel2020artificial} and one can come across many rare driving scenarios \cite{makansi2021exposing}. Newly produced cars are equipped with an increasing number of sensors, such as a driver monitoring and eye-gaze tracking system, to enable the development of more advanced driving support. An example of such an ADAS is driver intention recognition (DIR), which concerns identifying the driving maneuver aspirations of an ego-vehicle driver. Timely recognizing driver intentions {in traffic} enables proactive adaptation of the driver behavior. This can be important to mitigate potential traffic conflicts to promote cooperative resolutions for intricate traffic-related problem scenarios. Typically, DIR concerns recognizing the intention a few seconds before the actual maneuver \cite{fang2022behavioral}. Previous DIR studies predominantly rely on deep neural networks (DNNs) \cite{fang2022behavioral,vellenga2022driver}, but commonly neglect or lack explicit motivation for the balance between the model architecture complexity and the performance. Therefore, this paper primarily focuses on the impact of a DNN architecture on DIR performance, as previous studies have not empirically assessed this aspect, and neither have they considered the computational limitations of a car.}

For recognition tasks, it is common to combine data from different sensors, also referred to as different modalities. A key aspect of fusing multiple modalities is that every modality adds information that cannot be observed by any of the other sensors in the setup \cite{lahat2015multimodal}. DNNs are capable to learn from observations from multiple modalities in different ways \cite{azcarate2021data}, but their effectiveness depends, among others, on the context of the learning task, the quality of the data \cite{gudivada2017data}, and sometimes on the amount of data to learn from \cite{zhu2016we}.

To learn a task from multiple sensor observations, it is unclear upfront how complex an architecture should be. Breiman (2001)\cite{breiman2001statistical} notes that there can be many accurate models in a machine learning (ML) context, also referred to as the Rashomon effect. In an environment with limit computation capabilities {(e.g., in a car or truck)}, it can be critical to explore if reducing the complexity of an architecture affects the performance \cite{semenova2022existence}. Moreover, according to the universal approximation theorem, a single layer network can learn everything \cite{hinton2006fast}, but it does not specify how difficult the training of the model will be. Similarly, Heaton (2008) \cite{heaton2008introduction} argued that networks with two hidden layers can represent any arbitrary decision boundary. Next to the depth of a network, the width of a layer also needs to be determined. However, deciding how many hidden units are required solely based on the input and output dimensions of the data is difficult, but does affect the ability of the network to learn patterns \cite{karsoliya2012approximating}. Several studies have proposed a `rule-of-thumb' to select the initial number of hidden units. For example, dividing the number of samples by the scaled sum of the input and output dimensions \cite{hagan1997neural}, computing the square root of the dot product of the input and output dimensions \cite{masters1993practical}, or multiplying the sum of the in-- and output shape by two-thirds \cite{heaton2017heaton}. While these approaches might help to pick a starting point, they do not offer any empirical motivation for the DNN architecture design.

The architecture design relies ones existing knowledge and preferences, which potentially prohibits the exploration of better architectures and neglects an empirical motivation of the setup \cite{ren2021comprehensive}. Neural architecture search (NAS) can be used to evaluate and explore the performance of multiple DNN architectures \cite{elsken2019neural}. NAS requires limited human intervention and explores a pre-defined search space to automatically design DNN architectures that maximize an objective, such as the performance. NAS has been successfully applied to various fields (e.g., object detection \cite{chen2019detnas}, semantic segmentation \cite{liu2019auto}, and machine translation \cite{fan2020searching}), but to the best of our knowledge not yet to {DIR. The DIR context is different from previous NAS contexts because it relies on assessing implicit cooperative intentions of a driver in a dynamic ambiguous environment \cite{dietrich2019implicit,miller2022implicit} with limited computational capabilities.}

Previous DIR studies show similarities in DNN architectures. For example, Xing et al. (2020) \cite{xing2020ensemble} used a model consisting of a long short-term memory (LSTM) layer with 120 units, followed by a fully-connected (FC) layer with 100 units. Jain et al. (2016) \cite{Jain2016} implemented a single layer LSTM model with 64 units, Zyner et al. (2018) \cite{zyner2018recurrent} used three LSTM layers with 512 units each and a FC layers with 256 units, Guo et al. (2021) \cite{guo2021driver} utilized an LSTM architecture with 125 units, and Benterki et al. (2021) \cite{benterki2020artificial} used three LSTM layers with 256 units, and one LSTM layer with 128 units. While there is an overlap in the used architectures in previous studies, there is no explicit motivation for the types of layers, the number of hidden units or the number of layers. {Therefore, the main contribution of this article is an evaluation of eight search strategies for two DIR datasets. We analyse the top performing architectures for each search strategy and compare the influence of different combinations of the data and the architecture complexity on the intention recognition task. We use the Brain4Cars \cite{Jain2016} dataset for lane-change and turn maneuver DIR and the Five Roundabouts \cite{zyner2019acfr} to recognize turn maneuvers drivers intent to perform before entering a roundabout.} 

\section{Preliminaries}\label{sec:2}
\subsection{Information fusion}
{DIR studies use, among others, observations from vehicle dynamics sensors (e.g., velocity, yaw-rate), the driver monitoring system (e.g., head pose estimation, gaze estimation), or driving-scene observations (e.g., road user detection, lane detection, or traffic sign detection) as inputs to a DNN to recognize highway lane change or intersection turn maneuver intentions \cite{guo2021driver, rong2020driver, xing2020ensemble}.The relations between these observations can reveal important patterns that allow one to understand and recognize what a road user intends to do. However, making use of information from different sensors at different sampling rates is difficult, because cross-modality relations between low-level features are highly non-linear \cite{radu2018multimodal}. The main challenge is to find joint representations of the data that together yield an increased performance \cite{baltruvsaitis2018multimodal, meng2020survey}. 

\begin{figure*}[t]
    \centering
    \includegraphics[height=6.5cm, keepaspectratio,]{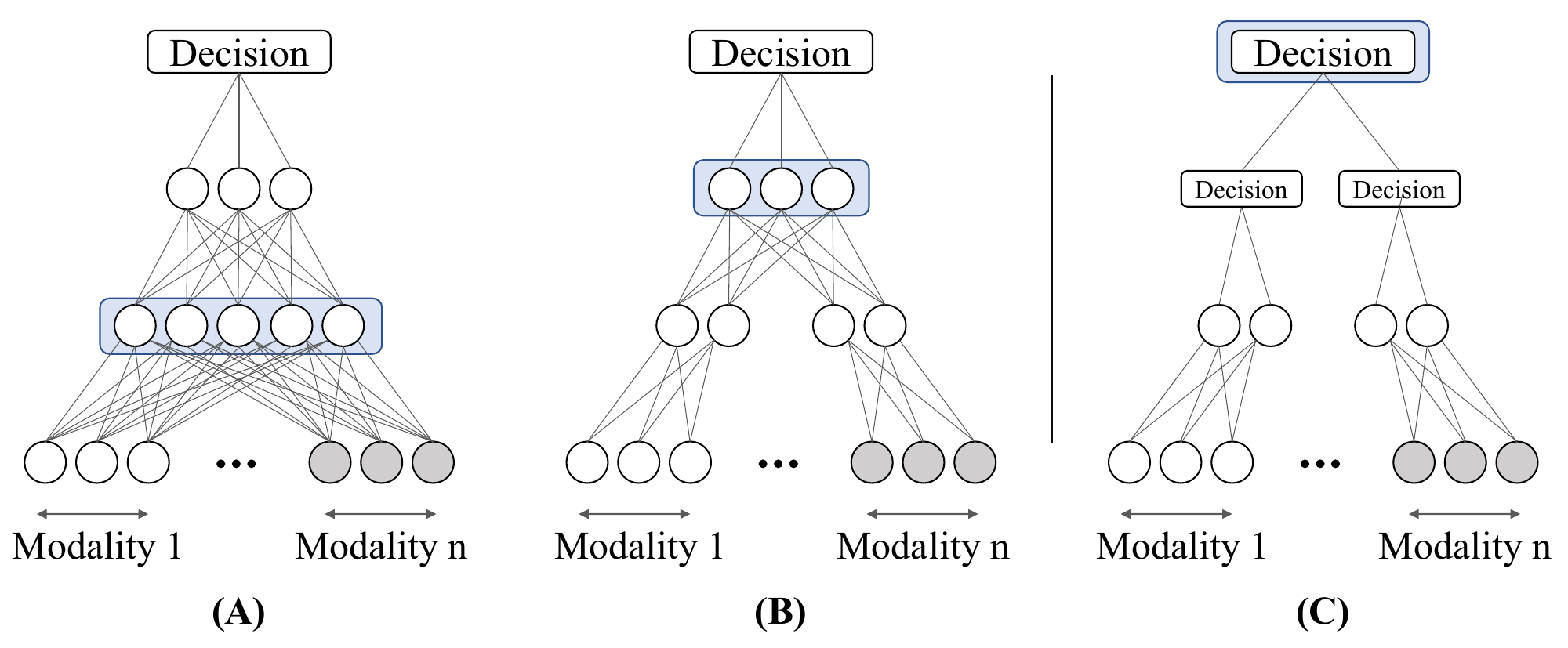}
   \caption{Schematic overview of deep learning fusion strategies, the blue box indicates where the fusion operation is performed. (A) The early fusion strategy expects a concatenated input vector of all modalities. (B) The intermediate fusion strategy first learns a representation for one or multiple modalities to learn a joint representation later in the network. (C) Late fusion first predicts per modality and combines all uni-modal predictions into a final decision.}\label{fig:fusion_strategies}
     \vspace{-0.0cm}
\end{figure*}

Typically fusion strategies for DNNs are categorized into: early, intermediate, or late fusion \cite{ramachandram2017deep} (see figure \ref{fig:fusion_strategies} for a schematic overview). `Early fusion' requires the DNN to learn cross-modal relations and patterns based on the raw low-level features. Relationships between modalities that require a higher level of abstraction might not be learned, and the strategy is sensitive to different sampling rates and commonly concatenates feature vectors of two modalities \cite{ramachandram2017deep, stahlschmidt2022multimodal}. `Intermediate fusion' strategies first learn a representation based on the raw input data and fuse the representations at a later stage in the same network. Instead of concatenating the learned representations, one can combine the learned representations in different ways. 
`Late fusion' combines uni-modal predictions into a final verdict. This does not allow for learning any interaction between the different modalities, but the uni-modal sub-networks of the DNN can learn specific representations per modality. 

\begin{figure}[t]
   \vspace{-0.1cm}
    \centering
    \includegraphics[height=6cm, keepaspectratio,]{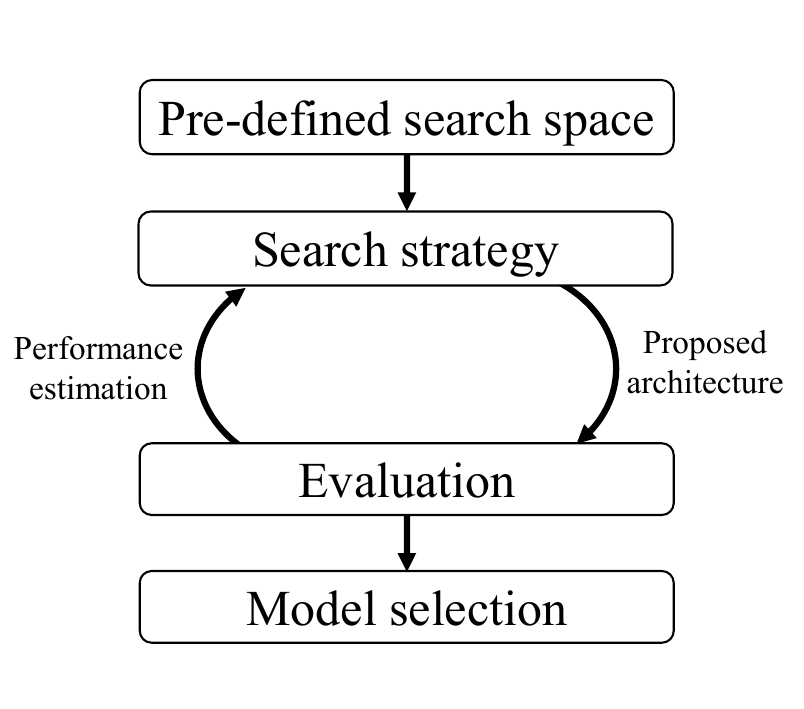}
   \vspace{-0.4cm}
   \caption{Overview of the neural architecture search framework. First, a search space has to be defined, after which a search strategy will sample DNN architectures. The estimated performance of the architecture is stored and the results are used by the search strategy as input to propose the next architecture. Figure adapted from \cite{elsken2019neural}.}\label{fig:NAS_strategies}
     \vspace{-0.0cm}
\end{figure}

\subsection{Neural architecture search}
Despite advancements of onboard computing platforms in cars, the computational resources are still quite limited, which makes it crucial to investigate whether reducing the complexity of a DNN yields comparable performance \cite{semenova2022existence}. To find the best model for a specific driving context and task, one needs to balance performance and complexity. Factors that affect the complexity of a model include the number of units or neurons in each layer, the number of hidden layers, the types of layers used, and how data from different sensors are combined. While deeper networks can perform better and allow for more abstract representations \cite{mhaskar2017and}, it is also important to consider the width of hidden layers, as it can affect the loss of information in a forward pass \cite{lu2017expressive}.

{Neural architecture search \cite{elsken2019neural,ren2021comprehensive} (NAS, see Figure \ref{fig:NAS_strategies}) is a framework to automatically test and sample DNN architectures based on a pre-defined search space, search and evaluation strategy. In a simple scenario, a network with a single branch without complex layer types, the block-encoding search space is parameterized by the number of layers, the operation types per layer, and the number of parameters associated with an operation (the number of units).  }

Exhaustively testing every possible network configuration of the search space is obviously computationally intensive. Therefore, a search strategy is used to find the optimal set of parameters given a pre-defined number of trials.  The search strategy samples an architecture within the boundaries of the defined search space to create, train, and evaluate that architecture. For all search strategies, the objective is to optimize a function $f$ given a set of parameters $(\theta_1, \theta_2, \dots, \theta_n)$. The objective function, is commonly referred to as a `black box' (i.e., a machine learning model that does not explain how predictions are made in a way a human can understand) because, the only thing that is known about the function is whether the objective is to find a minimum or maximum. 

\subsection{Search strategies}
 In this paper, we focus on multi-trial search strategies (see Table \ref{tab:archs} for the included strategies), because we are not considering the operations within a layer, but focus on optimizing the depth and width of existing DNN layers that can handle sequential data. Multi-trial strategies sample, train and evaluate a new architecture and update the sampling based on the previous evaluations. Random Search (RS) is a technique that randomly samples combinations from a search space with the objective of discovering the optimal architecture \cite{bergstra2012random}. The RS approach can serve as a baseline for comparing other search strategies, because it does not include any prior information to sample the next most promising architecture. Ideally, search strategies that include prior information should perform similar or better for a similar number of trials and avoid less promising architectures. 

A simple way to include prior information is the Hill-climbing algorithm \cite{elsken2018simple}. The search starts with an initial compact architecture and during every iteration a collection of marginally modified networks are assessed. The most promising network is selected as the basis for the subsequent iteration. In this way, the optimization procedure improves the neural network's performance by leveraging past evaluations.

\begin{table*}
\color{black}
  \footnotesize
    \centering
    \caption{Overview of the included search strategies for sampling DNN architectures from the pre-defined search space.\\}\label{tab:archs}
    \begin{tabular}{l|l}
        \toprule
      Name &   Basic intuition\\
        \midrule
      \begin{tabular}[l]{@{}l@{}} Random Search \cite{bergstra2012random}  \\
      \end{tabular}  &  \begin{tabular}[l]{@{}l@{}} 
     Samples random combinations from the \\ search space to find the best architecture   \end{tabular}   \\
           \midrule
 \begin{tabular}[l]{@{}l@{}} Hill-climbing \cite{elsken2018simple} 
      \end{tabular}  &  \begin{tabular}[l]{@{}l@{}} 
    Starts with a small initial network and \\ iteratively alters, evaluates the best performing \\ architectures. \end{tabular}   \\
         

         \midrule

      \begin{tabular}[l]{@{}l@{}} 
         Particle Swarm Optimization \cite{kennedy1995particle}  \end{tabular} 
          &  \begin{tabular}[l]{@{}l@{}} 
      Uses a number of particles (agents) to  \\ 
      iteratively navigate around the search space  \\
      guided toward the best know global position. \end{tabular}  \\
        \midrule
        \begin{tabular}[l]{@{}l@{}} 
          Regularized Evolution \cite{real2019regularized} 
          \end{tabular} 
          &  \begin{tabular}[l]{@{}l@{}} 
      Aging evolution algorithm that discards \\ 
      the oldest models in the population to \\
      allow for a wider search space exploration. \end{tabular}   \\

        \midrule
        \begin{tabular}[l]{@{}l@{}} Gaussian Process  \cite{li2020gp}
        \end{tabular}   &  \begin{tabular}[l]{@{}l@{}} 
      Employs a Gaussian Process to model \\ and update beliefs about the search space \\ to select the next architecture to evaluate.  \end{tabular}   \\

         \midrule
        \begin{tabular}[l]{@{}l@{}} Tree-structured Parzen \\ estmator (TPE) \cite{bergstra2011algorithms, bergstra2013making} \end{tabular}   &  \begin{tabular}[l]{@{}l@{}} 
      Uses a hierarchical Process based on a \\ performance threshold to generate the next \\ best architecture.  \end{tabular}   \\

         \midrule
   
        \begin{tabular}[l]{@{}l@{}} Policy based Reinforcement \\ Learning  \cite{zoph2016neural} \end{tabular}   &  \begin{tabular}[l]{@{}l@{}} 
      Leverages RL to train a RNN to generate  \\  architectures that yield the maximum  \\  expected performance. \end{tabular}   \\
       
        \midrule
        \begin{tabular}[l]{@{}l@{}}  Latent Action Neural Architecture \\ Search (LaNAS) \cite{wang2021sample} \end{tabular}   &  \begin{tabular}[l]{@{}l@{}} 
      Recursively partitions the search space into \\ a hierarchical tree structure to evaluate \\ to potential of future architectures. \end{tabular}   \\

        \bottomrule
    \end{tabular}
\end{table*}

Bayesian optimization (BO) search strategies incorporate a probabilistic belief on the objective function (called the surrogate function) and use an acquisition function to determine which architecture most likely improves the objective \cite{brochu2010tutorial}. The degree of exploration or greediness in selecting architectures varies depending on the acquisition function.  The posterior (refer to equation \ref{eq:bo}) helps to measure our knowledge about the objective function $f$. When we begin the search procedure, no architectures have been evaluated yet, so we start with limited knowledge about the objective function. As the search continues, the posterior is updated, which guides future sampling in an informed manner. In sum, the Bayesian optimization (BO) search algorithm involves repeatedly selecting an architecture by optimizing the acquisition function, evaluating the sampled architecture, and updating our beliefs and surrogate function until we meet the stopping criteria.

To perform BO one needs to decide how to summarize the joint conditional probabilities of the search space (the surrogate function). One probabilistic model that is often used for modeling the probabilistic beliefs is about the objective function is a Gaussian Process (GP) \cite{williams2006gaussian}  (refer to equation \ref{eq:GP}). In simple terms, a GP can be thought of as a collection of random variables. These variables have the property that any subset of them follows a joint multivariate Gaussian distribution. When $f$ is unknown, we assume that any set of evaluations of $\{\theta_1, \dots, \theta_n\}$ is distributed according to a multivariate Gaussian distribution, which gives an implied distribution for all inputs \cite{lindholm2019an}. The multivariate Gaussian distribution consists of a mean with the expected value of the distribution and a covariance matrix that consists of the variance and correlation between the variables. All values $k(x_i, x_j)$ of the covariance matrix $K$ are defined by a kernel function. If the points $x_i$ and $x_j$ are similar according to the kernel, then the output $y_i$ and $y_j$ are also expected to be similar. 

\begin{equation}
P(f|D) = P(D|f) \cdot P(f)
\label{eq:bo}
\end{equation}
Where:
\begin{conditions} 
P(f|D)  & = & posterior for $f$ given the evaluated samples D \\
f(\theta_i)  & = & objective function that returns the performance of  a sample $\theta_i$ \\
 \theta_i & = & specific architecture constellation from the search space \\
D & = & samples and outcomes ($\theta_i, f(\theta_i)$) that define the prior\\ 
& &  (updated every iteration)\\
\end{conditions}

\begin{equation}
     f \sim \mathcal{GP}(\mu, K)
     \label{eq:GP}
\end{equation}
Where:
\begin{conditions}
f(\theta_i)   & = & objective function that returns the outcome of  a sample $\theta_i$ \\
\mathcal{GP}(\mu, K) & = & Gaussian Process with mean function $\mu$ and covariance \\
&& kernel $K$
\end{conditions}

{An alternative BO approach is the tree-structured Parzen estmator (TPE) \cite{bergstra2011algorithms, bergstra2013making}}. TPE creates two hierarchical processes where {the evaluated samples $D$ are} split into two groups according to performance threshold $y^*$ (see equation \ref{eq:TPE}). For our example the objective would be to minimize $y$. The threshold can, for example, be based on performance of the sampled architectures. The sampled architectures are divided into two groups, one group with the top performing quantile, and the second group with the remaining evaluated architectures. Accordingly, the two hierarchical processes, $l(\theta)$ and $g(\theta)$ serve as models to generate the architecture variables. $l(\theta)$ is based on the samples that yields a performance lower than $y^*$, and $g(\theta)$ the samples of which the evaluation is higher than the threshold. The next architecture $\theta$ is then drawn from the model that satisfies the objective (in this example lowering $y$).

\begin{equation}
p(\theta|y)= 
\left\{\begin{array}{lr}
    l(\theta) & \text{if} \: y 	< y^*\\
     g(\theta) & \text{if} \:  y \geq  y^*\\ 
\end{array}
\right.
\label{eq:TPE}
\end{equation}

Where:
\begin{conditions} 
y^* & = & threshold value \\ 
l(\theta)  & = & distribution for $\theta$ when $y=f(\theta) < y^*$ \\
g(\theta)  & = & distribution for $\theta$ when $y=f(\theta) \geq y^*$ \\
\end{conditions}

Alternatively, the NAS problem can also be tackled using evolutionary algorithms (EA). Evolutionary computing (EC) or EAs are a group of algorithms that draw inspiration from Darwinian evolution \cite{darwish2020survey, fogel1995phenotypes} (see figure \ref{fig:EA}). The process starts with an initial population, which is a random set of architectures sampled from the search space. In each iteration, new individuals (architectures) are generated from existing parents using two types of operations, known as mutations. In a NAS context, mutations mean changing the structure of the architecture (e.g., removing a layer or changing the number of units). The selection of which architectures to consider for further refinement is based on the evaluation performance. The Regularized evolution search strategy \cite{real2019regularized} combines mutation and selection to iteratively evolve the neural network architectures, while incorporating a regularization term to balance exploration and exploitation during the search process. Another EC search strategy is particle swarm optimization (PSO) \cite{kennedy1995particle}. PSO is an algorithm inspired by the social behavior of animals like bird flocking. In the context of NAS, each potential neural network architecture is viewed as a particle with a certain velocity flying through the search space. The combination of a particles' own historical performance and the performance of other architectures in the swarm determines its next movement to gradually approach the optimal architecture.

\begin{figure}[t]
  \vspace{0.0cm}
    \centering
    \includegraphics[height=6.5cm, keepaspectratio,]{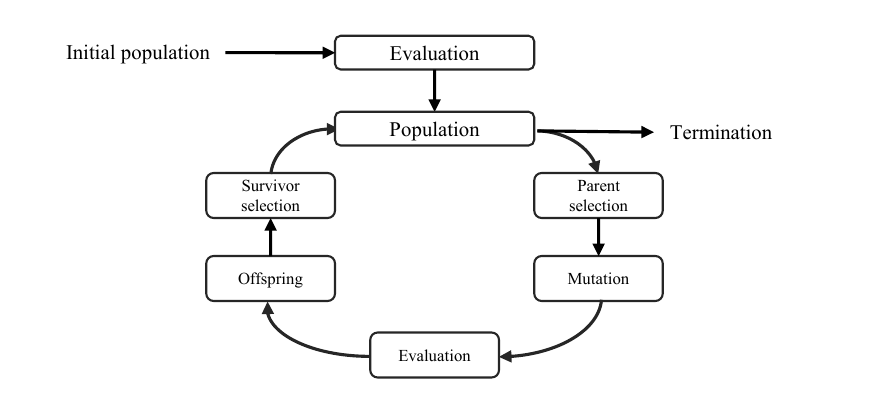}
  \vspace{-0.5cm}
  \caption{Generic framework describing evolutionary algorithms. The population is a set of architectures sampled from the search space in a NAS context, and every iteration is called a generation (figure adapted from \cite{darwish2020survey}).}\label{fig:EA}
     \vspace{-0.0cm}
\end{figure}  

Lastly, NAS can be framed as a reinforcement learning (RL) problem, where the agent's action corresponds to designing a network and the action space represents the search space. In RL, the agent's objective is to select an action $a_t$ that maximizes the future reward $r_{t+1}$, which represents the model's performance or an optimization objective. Each action modifies the environment, resulting in an updated state $s_{t+1}$ that provides information to the agent about the impact of the action. The agent uses the history $h_t = (\langle a_1,s_1,r_1 \rangle, \dots, \langle a_t, s_t,r_t \rangle)$ to select a new action (a new architecture to evaluate) that maximizes the future reward until the search process is complete. Policy-based RL approaches iteratively train a recurrent neural network (RNN) to improve the proposed architectures continuously \cite{zoph2016neural}. Wang et al. (2019) \cite{wang2019alphax} employed Monte-Carlo Tree Search (MCTS) to perform NAS. Intuitively, MCTS gradually constructs a tree and optimizes a policy that balances between exploration and exploitation \cite{browne2012survey}).  

The child-selection policy is repeatedly executed until a leaf node, a node without any children, is reached (step 1). The selection process uses the Upper Confidence Bounds (UCB1 \cite{auer2002finite,kocsis2006bandit}, see equation \ref{eq:UCB1}), where the UCB1 value is calculated for each node, and the child node with the highest value is selected until a leaf node is reached. If the leaf node is not a terminal node, a node that ends the run, one or more child nodes are created based on the available actions and the first node of the new nodes is selected (step 2). From this first new node M, a simulated roll-out is executed (step 3) until a terminal state is reached. The result is then added to the value ($V_i$) of all nodes from M to the root node, and the visit count ($n_i$) is updated at each node (step 4) \cite{baier2010power,chaslot2008monte}. Wang et al. (2021) \cite{wang2021sample} improved the efficiency of MCTS for NAS by recursively partitioning the search space into a hierarchical tree structure where some leaves have higher potential of yielding better performance than others.

\begin{equation}
UCB1 = v_i + C \cdot \sqrt{\frac{\ln N}{n_i}}
\label{eq:UCB1}
\end{equation}
Where:
\begin{conditions} 
v_i & = & value at state $S_i$\\
C & = & tuneable constant hyperparameter\\
n_i & = & number of visits at state $S_i$\\
N & = & total number of parent nodes of $S_i$\\
\end{conditions}

\section{Methods and metrics}\label{sec:3}
In this section we outline the used DNN layers that are able to handle sequential data and the evaluation metrics. The evaluation metrics is essential for NAS because it influences the search direction. Given the limited computational capabilities available in a car and to assess the relation between complexity and performance,  we consider the number of operations that an architecture has to perform to make a prediction.

\subsection{{Neural network layer types}}{We consider three DNN layer types to handle the sequential input data: a long-short term memory (LSTM), \cite{hochreiter1997long}, a temporal convolutional network (TCN) \cite{bai2018empirical}, and a time-series Transformers (TST) \cite{vaswani2017attention, zerveas2021transformer}). For each of the these layer types we provide a brief intuition.}

{\textbf{Long-short term memory.} To reuse information from previous time steps, a neural network requires some form of memory to store information. An LSTM \cite{hochreiter1997long} selectively remembers and forgets information over time by using memory cells that store information from previous time steps and uses gates to control the information flow. Upon receiving new inputs, the LSTM architecture decides which information to forget from the previous time step and which information to store in the memory cells \cite{sherstinsky2020fundamentals}.}

{\textbf{Temporal convolutional network.} A convolutional layer is designed to process structured grid data to perform an operation by sliding over the grid by using kernels and element-wise multiplications \cite{lecun1998gradient}. Where convolutional neural netwokrs (CNNs) are primarily used for visual data, Bai et al. (2018) \cite{bai2018empirical} introduced the TCN architecture to enable a CNN with dilated kernels and causal convolutions to process time series or sequential data to capture temporal dependencies. A dilated kernel, a convolutional kernel with spaces or gaps between its elements, allows a network to have a larger receptive field to capture information from a broader context (e.g., long-range temporal dependencies in time series data). A causal convolution ensures that only elements at time step $t$ and earlier are convolved to avoid information leakage.}

{\textbf{Time-series transformer.} Tokenization, positional encoding and the self-attention (SA) are the crucial components of the Transformer \cite{vaswani2017attention} architecture. The tokenization process is responsible for transforming the raw data into sequence representations. Subsequently, the positional encoding includes information about the location of each token in the sequence. The SA operation quantifies how much relevant information the other tokens in a sequence contain for a considered token. To enable the Transformer to handle time series data, Zerveas et al. (2021) \cite{zerveas2021transformer} introduced the TST, which uses a linear projection to transform the sequential input data to a vector to which the positional encoding is added.}

\subsection{Performance metrics.} A proper scoring metric should only reach its optimal value if the prediction confidence is equal to the ground truth according to Dawid \cite{dawid1986probability} and Gneiting and Raftery \cite{gneiting2007strictly}. In practical terms this means that metrics that use a threshold to determine the class do not qualify as a proper scoring metric. Suppose we have the following predictions and ground truth for a scenario with three classes: 
\[ \hat{y}_1: \left( \begin{array}{ccc} 0.1 & 0.6 & 0.3 \end{array} \right)\] 
\[ \hat{y}_2: \left( \begin{array}{ccc} 0.2 & 0.5 & 0.3 \end{array} \right)\] 
\[ ground\:truth: \left( \begin{array}{ccc} 0.0 & 1.0 & 0.0 \end{array} \right)\] 

A metric that uses a threshold of 0.5 to determine the class (e.g., accuracy or precision) would still produce the same prediction for both $\hat{y}_1$ and $\hat{y}_2$. For both instances the second class is higher or equal to the set threshold. If we assume that the second class is the correct answer and use a proper scoring rule instead, for example, the cross-entropy$\downarrow$\footnote{The arrows ($\uparrow/\downarrow$) indicate for every metric if higher or lower indicates a better result.}(CE, equation \ref{eq:ce}), the value would be 0.323 for $\hat{y}_1$, and 0.424 for $\hat{y}_2$. The CE is worse for $\hat{y}_2$ because the prediction for the correct class is less certain. Given this property of CE, we use it to guide the NAS performance evaluation. Complementary, we report the standard performance metrics (accuracy $\uparrow$, and the macro average for the precision $\uparrow$, recall $\uparrow$, and the f1-score $\uparrow$) to provide a broader evaluation perspective. 

\begin{equation}
CE(y_i, \hat{y}_i) = \frac{1}{N} \sum_{i=1}^N y_{i} \cdot log(\hat{y}_i)
\label{eq:ce}
\end{equation}
Where:
\begin{conditions} 
N  & = & number of predictions\\
i   & = & the $i^{th}$ instance from 1 to $N$\\
y_i  & = & ground truth for instance $i$\\
\hat{y}_i  & = & predicted confidence for instance $i$\\
\end{conditions}

\subsection{Model complexity.} 
Given that the computing resources in cars are shared between multiple safety related functionalities and are limited, it is important to consider the influence of model complexity on performance. A simplistic approach to assess model complexity, such as counting the number of trainable parameters, would ignore the impact of recurrent computations and only inform us about the storage space of the model \cite{laredo2020automatic}. { Hence, we adopt the real multiplications from Freire et al. (2022) \cite{freire2021performance} and estimate the floating-point operations (FLOPs) per architecture as an indication of the required computational complexity.}

\section{Experimental design}\label{sec:4}
\subsection{Problem formulation}

Suppose a driver that forms an intention to perform a driving maneuver (e.g., lane change or turn) $Y \in \{y_1,\dots,y_n\}$. The learning task is to recognize the driving intention $Y$ based on the observations $X_{i,t}$ over the time span of $k$ time steps.

\subsection{Datasets} We test the search strategies on an ego-vehicle driving maneuver and a roundabout DIR dataset. Refer to Figure \ref{fig:ruir_examples} for an overview of the included maneuvers.

\begin{figure}[t]
   \hspace*{-0.45cm}
    \centering
    \includegraphics[height=4.2cm, keepaspectratio,]{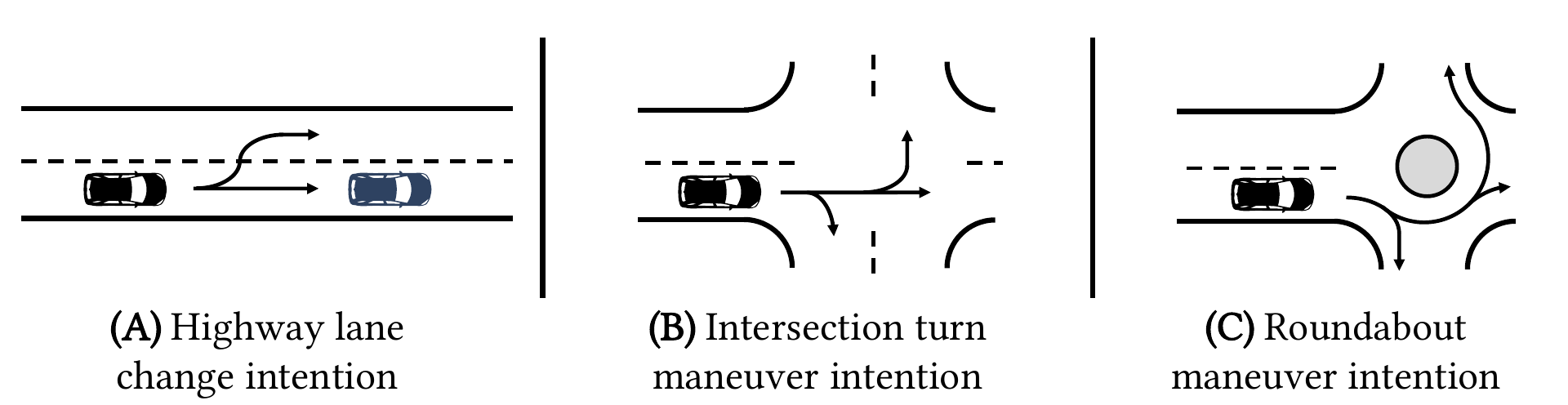}
   \vspace{-0.0cm}
   \caption{Overview of the driver intention recognition scenarios. (A). Ego-vehicle (black) driver's lane change intention \cite{Jain2016}. (B) Ego-vehicle driver's turn maneuver intention at an intersection \cite{Jain2016}. (C) Roundabout driving maneuver intention \cite{zyner2019acfr}.}\label{fig:ruir_examples}
     \vspace{-0.0cm}
\end{figure}  

\subsubsection{{Ego-vehicle driver intention recognition.}}

{The Brain4Cars dataset \cite{Jain2016} is an open-source driving dataset that consists of 124 left lane changes, 58 left turns, 123 right lane changes,  55 right turns, and 234 driving straight maneuvers with a five fold 80/20 train test split. The Brain4Cars dataset contains synchronized pre-processed data from in-vehicle sensors as well as sensors sensors outside of the vehicle for driving scene observations. The in-vehicle features consist of an aggregation of the video frames of tracked facial pixels in polar coordinates, and the 2D trajectories of facial landmark points. The driving-scene features consist of an indication of an upcoming intersection, the current lane of the ego-vehicle and the number of lanes to the left of the car. Lastly, the velocity of the car is the only vehicle dynamics feature that is provided in the dataset.}

\subsubsection{{Roundabout driving maneuver intention recognition.}}
{The Five Roundabout dataset \cite{zyner2019acfr} is an open-source naturalistic driving dataset collected at five roundabouts in Australia and consists of 5316 instances of vehicles taking the left turn, 16638 vehicles driving straight, and 1900 vehicles performing a right turn maneuver. We filtered out the 58 U-turn driving maneuvers, due to the class imbalance. For each car the distance and the relative angle to the roundabout entry and vehicle speed are collected at a rate of 25Hz. Contrary to Zyner et al. (2018) \cite{zyner2018recurrent}, we cut-off the observations 2 meters before the car enters the roundabout and do not include the origin direction of the vehicle.}

\begin{table}[t]
\footnotesize
\centering
\caption{Search space overview for both DIR datasets. The feed forward dimension and the number of attention heads parameters are only considered for the Time Series Transformer. \\}
\begin{tabular}{c|c|c}
\toprule
\textbf{Block} & \textbf{Parameter} & \textbf{Values} \\
\midrule
\multirow{7}{*}{\begin{tabular}[l]{@{}c@{}} Sequence \\ Representation \end{tabular}} & DNN layer type & LSTM, TCN, TST \\
\cline{2-3}
& Number of layers & 1  -- 4 \\
\cline{2-3}
& Number of units & 8 -- 256 \\
\cline{2-3}
& \begin{tabular}[l]{@{}c@{}}  Feed Forward \\ dimensions \textit{(TST only)} \end{tabular}  & 16 -- 256 \\
\cline{2-3}
& \begin{tabular}[l]{@{}c@{}} Number of attention \\  heads  \textit{(TST only)}  \\  \end{tabular}  & 2 -- 16 \\
\midrule
\multirow{2}{*}{\begin{tabular}[l]{@{}c@{}} Classification \\ Head \end{tabular}} & Number of layers & 1 -- 3 \\
\cline{2-3}
& Number of units & 8 -- 128 \\
\bottomrule
\end{tabular}\label{tab:sps}
\end{table}

\subsection{Experiments}
The goal of the experiment is to empirically motivate and evaluate DNN architectures for {DIR. We run NAS for a block-based encoding space that consists of a sequence layer and a classification head. Eight search strategies (Random Search, Gaussian Process, TPE, Hill-climbing, PSO, Regularized Evolution, Policy based RL, and La-NAS) run for 50 trials to sample combinations of different blocks with varying complexities. We evaluate the top performing architectures based on the CE performance per search strategy for each of the DNN layer types (LSTM, TCN and TST). To avoid that the search strategies only test a a single DNN layer type, we test the layer types separately for the different datasets for each search strategy. Additionally, we evaluate the effects of the early, intermediate and late fusion strategies for the in-cabin and exterior modalities in the Brain4Cars dataset. For the intermediate fusion approach, a sequence layer learns a representation per modality, after which the representations are concatenated and are used for the classification task. For the late fusion scenario, the uni-modal predictions are averaged.}

{Table \ref{tab:sps} represents the search space used for both DIR datasets. The DNN architecture is split up into two blocks, the sequence layers and the classification head to perform the recognition task.} {To avoid the influence of confounding factors on the performance, we keep the other tuneable hyperparameters stable where possible (a learning rate of $1e^{-4}$, a batch size of 256, dropout rate of 0.3, and the ADAM optimizer \cite{kingma2014adam}). For the Brain4Cars and FiveRoundabouts datasets, each search strategy runs for 50 trials for each sequence layer type, the architectures are trained for 100 epochs and with an early stopping patience of 25 epochs. After the search is completed, we perform the pre-defined five-fold cross validation for the top three architectures for each search strategy. The sampled architecture are automatically implemented in PyTorch (1.10) \cite{paszke2019pytorch} on an Ubuntu (20.04) server with a NVIDIA T4 GPU with 16GB of memory. We use the implementation from Wang et al. (2021) \cite{wang2021sample} for the LaNAS search strategy, NNI \cite{nni2021} for the regularized evolution and policy based RL search strategies, and Hyperactive \cite{hyperactive2021} for the remaining search strategies.}

\section{Results}\label{sec:5} 
In this section, we show the results for different search strategies compared to the original model designs, the influence of layer types (TST, TCN, LSTM) and the effect of the fusion strategies on the DIR performance for the Brain4Cars dataset.

\begin{table}[t]
    \caption{Overview of the original architecture performance compared to the best performing architecture sampled by the search strategies for the FiveRoundabout dataset.}
    \label{tab:frb_ss}
    \centering
    \small
    \begin{tabular}{@{}l@{\:}|l|@{\:}c@{\:}|@{\:}c@{\:}|@{\:}c@{\:}|@{\:}c@{\:}|@{\:}c@{\:}|@{\:}c@{}}
        \cmidrule[\heavyrulewidth]{2-8}
        & \textbf{Type}    & \textbf{CE ($\downarrow$)  }          & \textbf{Acc  ($\uparrow$)       }            & \textbf{Pr   ($\uparrow$)   }               &\textbf{ Re   ($\uparrow$)          }        & \textbf{F1            ($\uparrow$)       }  & \textbf{FLOPs $\cdot 10^{8}$} \\
        \midrule
        \cite{zyner2018recurrent} & LSTM & {0.334 $\pm $ 0.015}  & {88.63 $\pm $ 0.28 }     & 58.05 $\pm $ 1.10      & 55.05 $\pm $ 0.91      & 55.59 $\pm $ 0.77      & 1.314 \\
        \midrule
        RS      & TST    & 0.318 $\pm $ 0.021    & 88.67 $\pm $ 0.57      & 71.72 $\pm $ 2.58      & 66.02 $\pm $ 0.57      & {65.65 $\pm $ 2.19 }     & 0.300 \\
        
        HCO     & TST    & 0.314 $\pm $ 0.022    & 88.69 $\pm $ 0.85      & 80.51 $\pm $ 1.94      & 63.40 $\pm $ 0.85      & 63.92 $\pm $ 1.69      & 0.120 \\
        PSO     & LSTM    & 0.413 $\pm $ 0.040    & 86.04 $\pm $ 1.43      & 54.46 $\pm $ 1.70      & 51.63 $\pm $ 1.43      & 52.18 $\pm $ 1.66      & \textbf{0.011} \\
        Reg-Evo & TST   & 0.337 $ \pm $ 0.019 &  88.37 $ \pm $ 0.52 &  67.08 $ \pm $ 3.72 &  69.68 $ \pm $ 2.12 &   65.94 $ \pm $ 1.95 & 0.302  \\
        GP   & TST    & \textbf{0.313 $\pm $ 0.017}  & \textbf{88.94 $\pm $ 0.65 }     & 79.28 $\pm $ 2.38      & 63.41 $\pm $ 0.65      & 64.01 $\pm $ 2.51      & 0.124 \\
        TPE     & TST    & 0.322 $\pm $ 0.021    & 88.54 $\pm $ 0.72      & 74.26 $\pm $ 3.71      & 65.87 $\pm $ 0.72      & 63.83 $\pm $ 1.77      & 0.293 \\
        Pol-RL  & TST    & 0.318 $\pm $ 0.024    & 88.74 $\pm $ 0.72      & \textbf{81.85 $\pm $ 1.89  }    & 61.89 $\pm $ 0.72      & 62.48 $\pm $ 2.03      & 0.094 \\
         LaNAS    & TST    & 0.315 $\pm $ 0.012    & 88.75 $\pm $ 0.44      & 75.70 $\pm $ 2.43      & \textbf{66.30 $\pm $ 0.44 }     &\textbf{ 66.12 $\pm $ 2.25}      & 0.292 \\
         
        \bottomrule
    \end{tabular}

\end{table}

\begin{table}
 \caption{Overview of the original architecture performance compared to the best performing architecture sampled by the search strategies for the Brain4Cars dataset.}
    \label{tab:b4c_ss}
    \centering
   \color{black}
    \small
\begin{tabular}{@{}l@{\:}|l|@{\:}l@{\:}|@{\:}c@{\:}|@{\:}c@{\:}|@{\:}c@{\:}|@{\:}c@{\:}|@{\:}c@{\:}|@{\:}c@{}} 
 \cmidrule[\heavyrulewidth]{2-9}
& \textbf{Fusion} & \textbf{Type}    & \textbf{CE ($\downarrow$)  }          & \textbf{Acc  ($\uparrow$)       }            & \textbf{Pr   ($\uparrow$)   }               &\textbf{ Re   ($\uparrow$)          }        & \textbf{F1            ($\uparrow$)       }  & \textbf{FLOPs $\cdot 10^{8}$} \\ \midrule 

\cite{Jain2016} &  Inter & LSTM &  0.512 $ \pm $ 0.013 &  82.09 $ \pm $ 1.33 &  86.56 $ \pm $ 0.72 &  86.10 $ \pm $ 1.09 &  85.84 $ \pm $ 1.09 &  \textbf{0.003} \\
\midrule 
RS &  Inter & LSTM &  {0.437 $ \pm $ 0.016} &   85.93 $ \pm $ 1.60 &  89.33 $ \pm $ 1.04 &  89.03 $ \pm $ 1.25 &  88.84 $ \pm $ 1.23 &  0.076 \\ 
HCO &  Early & LSTM &  0.445 $ \pm $ 0.011 &  85.93 $ \pm $ 0.46 &  89.13 $ \pm $ 0.12 &  88.99 $ \pm $ 0.37 &  88.68 $ \pm $ 0.31 &  {0.016 }\\ 
PSO &  Early & LSTM &  0.453 $ \pm $ 0.022 &  {86.14 $ \pm $ 0.07} &  {89.34 $ \pm $ 0.34} &  {89.16 $ \pm $ 0.04 }&  {88.85 $ \pm $ 0.11} &  {0.016} \\  
Reg-Evo &  Inter & LSTM &  0.436 $ \pm $ 0.030 &  84.87 $ \pm $ 2.83 &  88.28 $ \pm $ 1.94 &  88.43 $ \pm $ 1.89 &  87.83 $ \pm $ 2.32 &   0.070\\ 
GP &  Early & LSTM &  0.459 $ \pm $ 0.033 &  85.29 $ \pm $ 1.37 &  88.53 $ \pm $ 1.21 &  88.55 $ \pm $ 1.03 &  88.22 $ \pm $ 1.07 &  {0.016} \\ 
TPE &  Inter & LSTM &  0.447 $ \pm $ 0.011 &   84.22 $ \pm $ 0.40 &  88.14 $ \pm $ 0.54 &  87.72 $ \pm $ 0.28 &  87.45 $ \pm $ 0.33 &  0.085 \\  
Pol-RL &  Inter & LSTM & \textbf{0.416 $ \pm $ 0.032 }&  \textbf{86.35 $ \pm $ 1.75} &    \textbf{89.40 $ \pm $ 1.70} &  \textbf{89.44 $ \pm $ 1.35} &  \textbf{89.19 $ \pm $ 1.56} &  0.125 \\ 
LaNAS &  Early & LSTM &  0.455 $ \pm $ 0.018 &  86.14 $ \pm $ 0.76 &  89.36 $ \pm $ 0.42 &  89.21 $ \pm $ 0.55 &  88.88 $ \pm $ 0.61 &  0.036 \\ \bottomrule

\end{tabular} 
   
\end{table}

\subsection{{Search strategies}}
For both the FiveRoundabouts and Brain4Cars datasets, we tested the architectures from the original studies for our training setup. For both DIR datasets, all search strategies, except for the PSO search strategy for the FiveRoundabouts dataset, sampled an architecture that performed better than the architecture from the original study. Tables \ref{tab:frb_ss} and \ref{tab:b4c_ss} show the top performing model for each search strategy.  For the FiveRoundabouts dataset (see Table \ref{tab:frb_ss}), the TST based architectures performed best. Although, the architecture from Zyner et al. (2018) \cite{zyner2018recurrent} performs relatively well from a CE and accuracy perspective, the TST clearly outperforms the LSTM architecture when considering the Precision, Recall and F1-score. Moreover, all architectures sampled by the search strategies demonstrated that equal performance can be achieved with a less complex model. See table \ref{tab:frb_top_models} for an overview of the top architectures for the FiveRoundabouts dataset. 

\begin{table}[t]
    \caption{Top three best performing architectures for the FiveRoundabouts dataset. ATT=Attention.}
    \centering
   \color{black}
    \small
    \begin{tabular}{l|cc}
        \toprule
         \textbf{No.}   & \textbf{Sequence layers}   & \textbf{Classification head} \\
        \midrule
        1 & 4 $\times$ TST layers, 119 sequence dim, 16 att heads, 256 dim  & 2 $\times$ Linear layers, 119 dim  \\ \midrule
           2 & 4 $\times$ TST layers, 124 sequence dim,  12 att heads, 128 att dim  & 2 $\times$ Linear layers, 159 dim  \\ \midrule
              3 & 4 $\times$ TST layers, 212 sequence dim ,2 att heads, 212 att dim  & 2 $\times$ Linear layers, 211 dim  \\
        \bottomrule
    \end{tabular}
    \label{tab:frb_top_models}
\end{table}

\begin{table}[t]
    \caption{Top three best performing architectures for the Brain4Cars dataset.}
    \centering
   \color{black}
    \small
    \begin{tabular}{l|l|cc}
        \toprule
         \textbf{No.}  & \textbf{Fusion} & \textbf{Sequence layers}   & \textbf{Classification head} \\
        \midrule
        1 & Intermediate &\begin{tabular}[l]{@{}c@{}} Modality 1: 4 $\times$ LSTM layers, 194 dim \\  Modality 2: 2 $\times$ LSTM layers, 202 dim \\ Shared: 1 $\times$ LSTM layers, 113 dim \end{tabular}  & 3 $\times$ Linear layers, 166 dim  \\ \midrule
   2 & Intermediate &\begin{tabular}[l]{@{}c@{}} Modality 1: 3 $\times$ LSTM layers, 222 dim \\  Modality 2: 1 $\times$ LSTM layers, 184 dim \\ Shared: 3 $\times$ LSTM layers, 180 dim \end{tabular}  & 2 $\times$ Linear layers, 88 dim  \\ \midrule
       3 & Intermediate &\begin{tabular}[l]{@{}c@{}} Modality 1: 3 $\times$ LSTM layers, 215 dim \\  Modality 2: 1 $\times$ LSTM layers, 181 dim \\ Shared: 3 $\times$ LSTM layers, 163 dim \end{tabular}  & 1 $\times$ Linear layers, 88 dim  \\
        \bottomrule
    \end{tabular}
    \label{tab:b4c_top_models}
\end{table}

The Brain4Cars performance results (Table \ref{tab:b4c_ss}) show that for all search strategies the LSTM architectures performed best. Similarly to the FiveRoundabouts results, the search strategies found a better performing architecture than the original architecture. However, a more complex architecture was required for improved performance. See table \ref{tab:b4c_top_models} for the top three architectures.

\subsection{{Model type}}

\begin{figure}[t]
\centering
\begin{subfigure}{0.49\textwidth}
\centering
\begin{tikzpicture}[scale=0.7]
\begin{axis}[
xlabel= {Floating-Point Operations},
ylabel={Average Validation Cross-entropy},
ymin=0.28,
ymax=0.55,
ytick={0.0, 0.1, 0.2, 0.3, 0.4, 0.5, 0.6, 0.7, 0.8, 0.9,1},
ymajorticks=true,
label style={font=\small},
tick label style={font=\small} ,
    legend pos=north east,
    ymajorgrids=true,
    grid style=dashed,
]

\pgfplotstableread[col sep=comma]{data/data_frb.csv}{\datatable}

\addplot[
scatter,
only marks,
scatter src=explicit symbolic,
scatter/classes={mtst={mark=square,blue, mark size=3}, tcn={mark=triangle,red, mark size=3}, lstm={mark=o,black, , mark size=3},original={mark=10-pointed star,cyan,mark size=4}},
] table [x=x, y=y, meta=label] {\datatable};
\legend{TST,TCN,LSTM,Original}
\end{axis}
\end{tikzpicture}
       \caption{FiveRoundabout Dataset}
   \end{subfigure}
\begin{subfigure}{0.50\textwidth}
\centering
\begin{tikzpicture}[scale=0.7]
\begin{axis}[
xlabel= {Floating-Point Operations},
ymin=0.35,
ymax=1.1,
ytick={0.0, 0.1, 0.2, 0.3, 0.4, 0.5, 0.6, 0.7, 0.8, 0.9,1},
ymajorticks=true,
label style={font=\small},
tick label style={font=\small} ,
    legend pos=south east,
    ymajorgrids=true,
    grid style=dashed,
]

\pgfplotstableread[col sep=comma]{data/data_b4c.csv}{\datatable}

\addplot[
scatter,
only marks,
scatter src=explicit symbolic,
scatter/classes={mtst={mark=square,blue, mark size=3}, tcn={mark=triangle,red, mark size=3}, lstm={mark=o,black, , mark size=3}, original={mark=10-pointed star,cyan,mark size=4}}
] table [x=x, y=y, meta=label] {\datatable};
\end{axis}
\end{tikzpicture}
       \caption{Brain4Cars Dataset}
   \end{subfigure}

\caption{Visualization of the five fold cross-validation performance and model complexity for the top 3 models for each type sampled by the search strategies. Best viewed in color.} 
\label{fig:dir_model_type}
\end{figure}
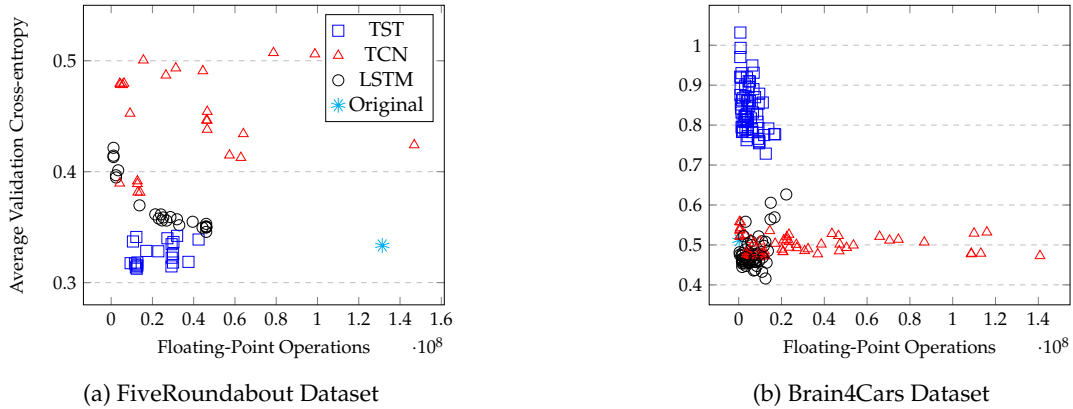

\begin{table}[t]
    \caption{Overview of the best performing architectures per model type for the FiveRoundabout dataset.}
    \centering
   \color{black}
    \small
    \begin{tabular}{l|@{\:}c@{\:}|@{\:}c@{\:}|@{\:}c@{\:}|@{\:}c@{\:}|@{\:}c@{\:}|@{\:}c@{}}
        \toprule
         \textbf{Type}   & \textbf{CE ($\downarrow$)  }          & \textbf{Acc  ($\uparrow$)       }            & \textbf{Pr   ($\uparrow$)   }               &\textbf{ Re   ($\uparrow$)          }        & \textbf{F1            ($\uparrow$)       }  & \textbf{FLOPs $\cdot 10^{8}$} \\
        \midrule
         TST    & \textbf{0.313 $\pm $ 0.017}  & \textbf{88.94 $\pm $ 0.65 }     & \textbf{79.28 $\pm $ 2.38}    & \textbf{63.41 $\pm $ 0.65 }     &\textbf{ 64.01 $\pm $ 2.51}      & \textbf{0.124} \\
         TCN    & 0.381 $\pm $ 0.018    & 88.20 $\pm $ 0.79      & 56.51 $\pm $ 0.85      & 55.53 $\pm $ 0.69      & {55.36 $\pm $ 0.88 }     & 0.140 \\
         LSTM    & 0.346 $\pm $ 0.018    & 88.06 $\pm $ 0.29      & 57.71 $\pm $ 1.26      & 54.08 $\pm $ 0.44      & {54.79 $\pm $ 0.38 }     & 0.461 \\
        \bottomrule
    \end{tabular}

    \label{tab:frb_model_type}
\end{table}

\begin{table}[t!]
    \caption{Overview of the best performing architectures per model type for the Brain4Cars dataset.}
    \centering
   \color{black}
    \small
    \begin{tabular}{l|l|@{\:}c@{\:}|@{\:}c@{\:}|@{\:}c@{\:}|@{\:}c@{\:}|@{\:}c@{\:}|@{\:}c@{}}
        \toprule
         \textbf{Type} & \textbf{Fusion}   & \textbf{CE ($\downarrow$)  }          & \textbf{Acc  ($\uparrow$)       }            & \textbf{Pr   ($\uparrow$)   }               &\textbf{ Re   ($\uparrow$)          }        & \textbf{F1            ($\uparrow$)       }  & \textbf{FLOPs $\cdot 10^{8}$}  \\
        \midrule
         TST    & Inter & 0.729 $ \pm $ 0.056  & 74.63 $ \pm $ 5.02   & 74.46 $ \pm $ 4.80 & 75.91 $ \pm $ 5.37 &   73.06 $ \pm $ 6.01 & 0.126 \\
         TCN    & Inter & 0.469 $ \pm$ 0.021 &    85.29 $ \pm$ 1.40 &   88.19 $ \pm$ 1.37 &   88.68 $ \pm$ 1.15 &   88.27 $ \pm$ 1.39 &  \textbf{0.098}  \\
         LSTM   & Inter & \textbf{0.416 $ \pm $ 0.032 }&  \textbf{86.35 $ \pm $ 1.75} &    \textbf{89.40 $ \pm $ 1.70} &  \textbf{89.44 $ \pm $ 1.35} &  \textbf{89.19 $ \pm $ 1.56} &  0.125 \\
        \bottomrule
    \end{tabular}

    \label{tab:b4c_model_type}
\end{table}

Figure \ref{fig:dir_model_type} visualizes the model performance in terms of five-fold cross-validation cross-entropy and the complexity in FLOPs for all top 3 performing layer types per search architecture. For the Brain4Cars dataset this includes three times more models, because each search strategy ran for each fusion strategy for each of the layer types. Tables \ref{tab:frb_model_type} and \ref{tab:b4c_model_type} contain the results for the best performing architectures per model type. Figure \ref{fig:dir_model_type} indicates that a more complex model in terms of FLOPs does not lead to improved performance. Moreover, for the fixed training setup, no model type clearly outperforms the other types for both of these DIR datasets. As shown in table \ref{tab:frb_model_type} for the FiveRoundabout dataset, the TST model performs best whereas for the Brain4Cars dataset both the TCN and LSTM models perform better (see Table \ref{tab:b4c_model_type}).

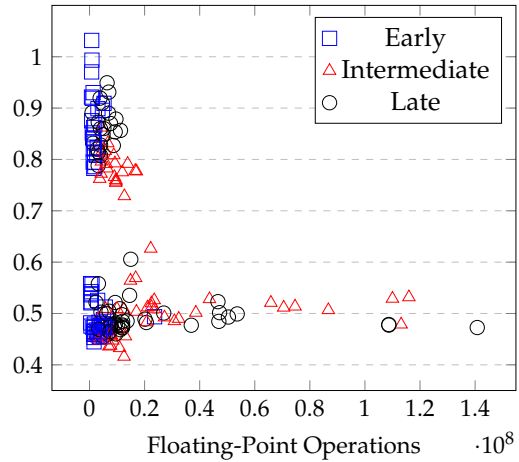
\begin{figure}[t]
\centering
\begin{tikzpicture}[scale=0.9]
\begin{axis}[
xlabel= {Floating-Point Operations},
ymin=0.35,
ymax=1.1,
ytick={0.0, 0.1, 0.2, 0.3, 0.4, 0.5, 0.6, 0.7, 0.8, 0.9,1},
ymajorticks=true,
label style={font=\small},
tick label style={font=\small} ,
    legend pos=north east,
    ymajorgrids=true,
    grid style=dashed,
]

\pgfplotstableread[col sep=comma]{data/data_b4c_fusion.csv}{\datatable}

\addplot[
scatter,
only marks,
scatter src=explicit symbolic,
scatter/classes={early={mark=square,blue, mark size=3}, inter={mark=triangle,red, mark size=3}, late={mark=o,black, , mark size=3}},
] table [x=x, y=y, meta=label] {\datatable};
\legend{Early,Intermediate, Late}
\end{axis}
\end{tikzpicture}

\caption{Overview of the fusion approach influence on the five-fold cross-validation model performance for the Brain4Cars dataset. The top three architectures per search strategy are included. Best viewed in color.}
\label{fig:b4c_fusion}
\end{figure}

\begin{table}[t]
    \caption{Overview of the best performing architectures per fusion approach for the Brain4Cars dataset.}
    \label{tab:b4c_fusion}
    \centering
   \color{black}
    \small
\begin{tabular}{l|l|@{\:}c@{\:}|@{\:}c@{\:}|@{\:}c@{\:}|@{\:}c@{\:}|@{\:}c@{\:}|@{\:}c@{}}
        \toprule
         \textbf{Fusion} & \textbf{Type}   & \textbf{CE ($\downarrow$)  }          & \textbf{Acc  ($\uparrow$)       }            & \textbf{Pr   ($\uparrow$)   }               &\textbf{ Re   ($\uparrow$)          }        & \textbf{F1            ($\uparrow$)       }  & \textbf{FLOPs $\cdot 10^{8}$}  \\
        \midrule
         Early    & LSTM & 0.445 $ \pm $ 0.011 &   85.93 $ \pm $ 0.46 &   89.13 $ \pm $ 0.12 &   88.99 $ \pm $ 0.37 &   88.68 $ \pm $ 0.31  &  0.164\\
         Inter    & LSTM & \textbf{0.416 $ \pm $ 0.032 }&  \textbf{86.35 $ \pm $ 1.75} &    \textbf{89.40 $ \pm $ 1.70} &  \textbf{89.44 $ \pm $ 1.35} &  \textbf{89.19 $ \pm $ 1.56} &  0.125 \\
         Late   & LSTM & 0.455 $ \pm $ 0.007 &   84.01 $ \pm $ 1.31 &   87.83 $ \pm $ 0.74 &   87.65 $ \pm $ 1.08 &   87.49 $ \pm $ 0.96 &  0.0497 \\
        \bottomrule
    \end{tabular}

\end{table}

\subsection{{Fusion strategy}}
Figure \ref{fig:b4c_fusion} and Table \ref{tab:b4c_fusion} display the fusion strategy influence for the Brain4Cars dataset. Figure \ref{fig:b4c_fusion} shows the same performance results as Figure \ref{fig:dir_model_type} (B), but indicates the fusion approach instead of the layer type. The early fusion architectures are all among the less complex models, whereas for the intermediate and late fusion there are both simple and complex architectures that yield similar performance. Table \ref{tab:b4c_fusion} shows that the best performing intermediate fusion architecture is slightly better than the best performing early or late fusion architecture. 

\section{Discussion}\label{sec:6}
The top performing architectures sampled by the search strategies do not clearly indicate that a particular search strategy leads to better architectures for the FiveRoundabouts dataset. Table \ref{tab:b4c_ss} shows for the Brain4Cars dataset that the Policy based RL search strategy sampled the best performing architecture for all performance evaluation metrics. Besides the search strategies, it is also dataset dependent which layer type performs best. For this experiment, we only included the model architecture parameters in the search space. Given this constraint, we can only observe that there are performance differences between the layer types per dataset, but there is no indication that a certain layer type performs best for both. The relationship between increased complexity and improved performance is not clear based on the results in Figure \ref{fig:dir_model_type}. There are plenty of factors, such as regularization, number of training epochs, the early stopping patience, optimizer type, weight decay, learning rate schedules, which can also affect the performance of an architecture. In this experiment the search was only performed on a single fold for both datasets, which influences the sampled architectures and eventually the performance over five folds. Thus, for this experimental setting, we did not observe a single architecture clearly outperforming all others, which hints to the Rashomon effect \cite{breiman2001random}. 

While NAS helps to explore and evaluate a landscape of options defined by the researcher, it does not consider unforeseen scenarios, such as failing sensors, that can potentially impact the performance when the car drives on the road. The performance evaluation relies one a single proper scoring rule, but the results of the FiveRoundabout DIR results show how this can be misleading. The original and the top performing architectures per search strategy perform roughly similar based on the CE and accuracy. In a safety-critical system, one can decide to prioritize high precision to minimize false alarms, or high recall to ensure that you capture as many relevant events as possible. The model with the lowest CE in Table \ref{tab:frb_ss} does not yield the highest precision, recall and F1 score. Thus, for DIR single metric based evaluation can be insufficient. For future studies, it could be interesting to consider the architecture's robustness (e.g., quantify the model uncertainty, test performance for the scenario with failing sensors, effects of context shifts, or the performance of the best architecture on a related dataset). Similarly for the complexity evaluation, additional metrics (e.g., power energy consumption, latency, throughput, and chip cost \cite{sze2017efficient}) can be considered to provide more relevant information for the actual integration of DIR methods \cite{lones2021avoid}. 

\section{Conclusion}\label{sec:7}
Deep neural networks (DNNs) have become the preferred approach for DIR studies \cite{vellenga2022driver}. For a constraint experimental setting, we evaluated eight search strategies for a block-based encoding search space with three different DNN layer types (long-short term memory [LSTM], temporal convolutional networks [TCN], and multivariate time-series Transformer [TST]) for two driver intention recognition (DIR) datasets. We observed a performance increase compared to the proposed architectures in the original DIR studies, but we did not observe an overall best performing search strategy. The sampled architectures did not indicate that a more complex architecture leads to a performance improvement. For the FiveRoundabout DIR dataset, the TST-based architecture performed best. For the Brain4Cars DIR dataset, an LSTM-based intermediate fusion architecture yielded the best results. Altogether, for a fixed training setup, we showed that performing NAS leads to performance improvements. Therefore, we conclude that explicitly empirically exploring and motivating the architecture of a DNN is an essential modeling step. 

\bibliographystyle{plainnat}



\end{document}